\documentclass{article}

\usepackage[preprint]{neurips_2025}

\usepackage[utf8]{inputenc}
\usepackage[T1]{fontenc}
\usepackage{graphicx}
\usepackage{float}
\usepackage{tablefootnote}
\usepackage{xcolor}
\usepackage{amssymb}
\usepackage{subcaption}
\usepackage{nicefrac}
\usepackage{microtype}
\usepackage{booktabs}
\usepackage{amsfonts}
\usepackage{url}
\usepackage[colorlinks=true,linkcolor=black,citecolor=black,urlcolor=blue]{hyperref}
\setcitestyle{numbers,square}

\title{CureAgent: A Training-Free Executor-Analyst Framework for Clinical Reasoning}

\author{%
Tingting Xie\thanks{Project Lead and Corresponding Author: \texttt{helloxietingting@gmail.com}. Orchestrated project resources; conceived the methodology; designed the Multi-Agent architecture; led the implementation and writing.} and Yixin Zhang\thanks{zyxcambridge@gmail.com. Contributed computational resources (GPUs); assisted with the initial deployment of the \texttt{txagent} environment; participated in experimental analysis on Gemini models.} \\
Team: CureAgent, prize-winning solution of \href{https://curebench.ai/}{CURE-Bench@NeurIPS 2025} \\
}

\begin{document}

\maketitle
\begin{abstract}
Current clinical agent built on small LLMs, such as TxAgent suffer from a \textit{Context Utilization Failure}, where models successfully retrieve biomedical evidence due to supervised finetuning but fail to ground their diagnosis in that information. In this work, we propose the Executor-Analyst Framework, a modular architecture that decouples the syntactic precision of tool execution from the semantic robustness of clinical reasoning. By orchestrating specialized TxAgents (Executors) with long-context foundation models (Analysts), we mitigate the reasoning deficits observed in monolithic models. Beyond simple modularity, we demonstrate that a Stratified Ensemble strategy significantly outperforms global pooling by preserving evidentiary diversity, effectively addressing the information bottleneck. Furthermore, our stress tests reveal critical scaling insights: (1) a \textit{Context-Performance Paradox}, where extending reasoning contexts beyond 12k tokens introduces noise that degrades accuracy; and (2) the \textit{Curse of Dimensionality} in action spaces, where expanding toolsets necessitates hierarchical retrieval strategies. Crucially, our approach underscores the potential of training-free architectural engineering, achieving state-of-the-art performance on CURE-Bench without the need for expensive end-to-end finetuning. This provides a scalable, agile foundation for the next generation of trustworthy AI-driven therapeutics. Code will be released on https://github.com/June01/CureAgent.
\end{abstract}

\section{Introduction} 

The advent of large language models (LLMs) such as Med-PaLM~\cite{singhal2023medpalm} and GPT-4~\cite{nori2023capabilities} has heightened anticipation for transformative advances in clinical decision support, given their capacity for broad knowledge and fluent reasoning. However, real-world medical reasoning requires more than just internal knowledge; it demands the ability to actively retrieve and synthesize information from constantly updating biomedical sources (e.g., FDA labels~\cite{kass2016openfda}, OpenTarget database~\cite{ochoa2023opentarget} and Human Phenotype Ontology~\cite{talapova2023human}). The CURE-Bench competition~\cite{curebench2025} addresses this critical gap by benchmarking agents not merely on static knowledge, but on their ability to utilize a vast ToolUniverse~\cite{gao2025tooluniverse} to perform structured, multi-step reasoning for clinically grounded tasks.

\begin{figure}[htbp]
  \begin{subfigure}{0.58\linewidth}
    \centering
    \includegraphics[width=\linewidth]{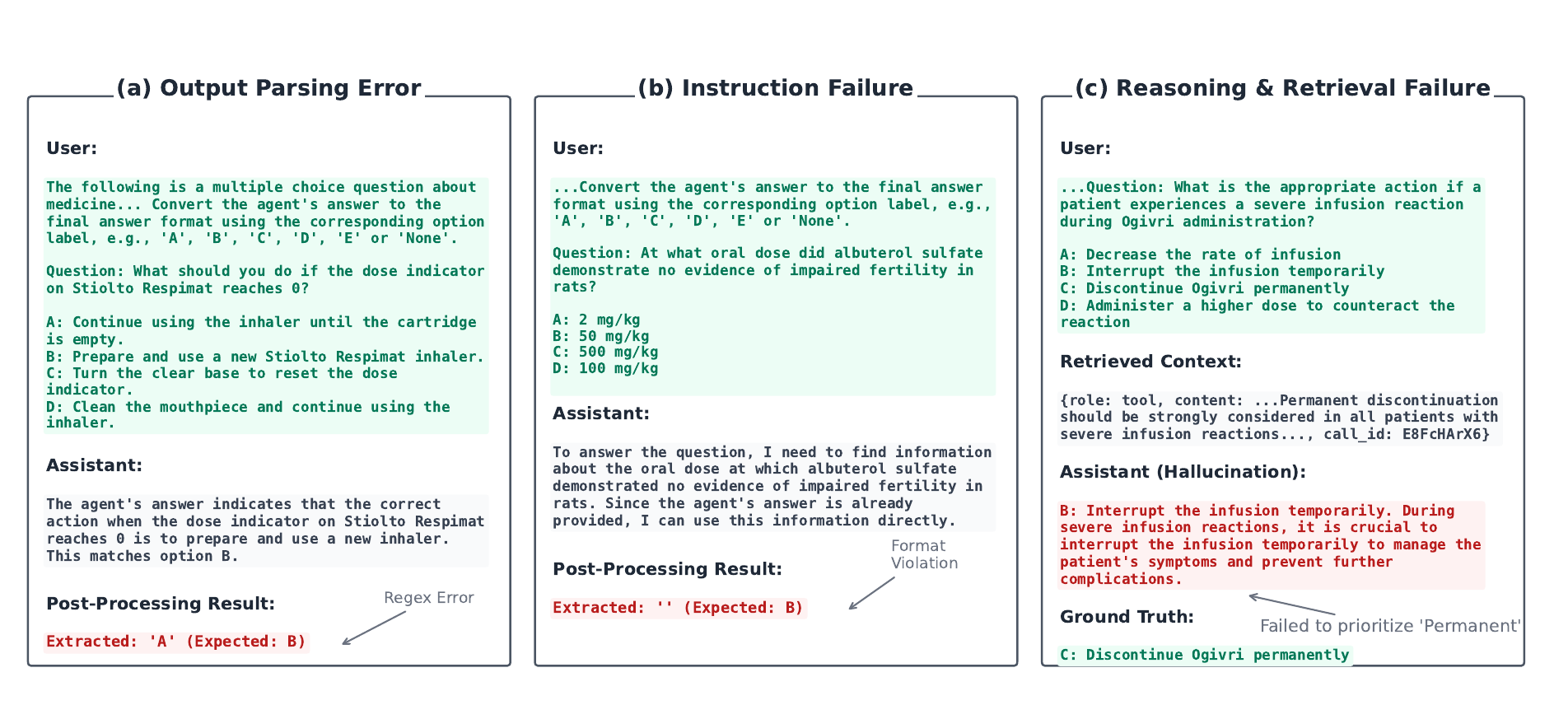}
  \end{subfigure}
  \hfill
  \centering
  \begin{subfigure}{0.4\linewidth}
    \centering
    \includegraphics[width=\linewidth]{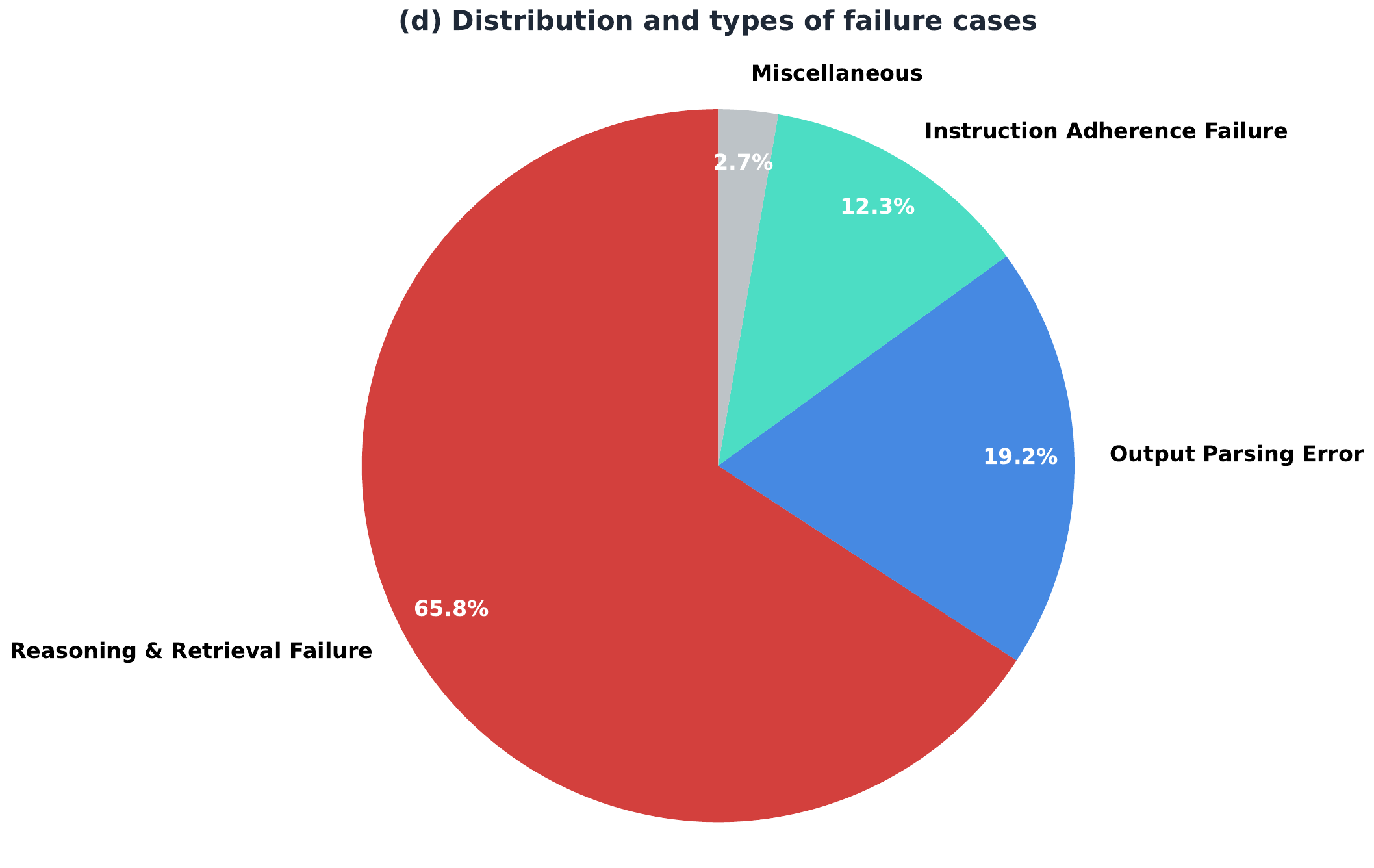}
  \end{subfigure}
  \caption{Analysis of TxAgent failed cases on the validation set. (a)(b)(c) Examples of failed cases highlighting different error modes in TxAgent. (d) Distribution and types of failure modes (\(n=73\) out of 413 multi-choice questions) observed in TxAgent, providing insights into common pitfalls and areas for improvement.}
  \label{fig:failure-case-analysis-combined}
\end{figure}

Specialized tool-augmented models like TxAgent~\cite{gao2025txagent} (the competition baseline) serve as a strong starting point. Fine-tuned specifically for this domain and curated for precise therapeutics, TxAgent demonstrates significant utility in executing complex API calls and multi-turn conversation to access biomedical knowledge and reasoning. However, a notable gap remains in achieving truly precise decision-making. Our analysis of failure cases with ground truth in the validation set (Figure~\ref{fig:failure-case-analysis-combined}(d)) reveals four primary error categories, with \textit{Reasoning \& Retrieval Failure} constituting the dominant share (65.8\%), while an example is shown in Figure~\ref{fig:failure-case-analysis-combined}(c). While this category includes instances of intrinsic model incapability or failures to invoke the correct tools, the vast majority represent a critical \textit{Context Utilization Failure}: the agent successfully gathers sufficient evidence but fails to ground its final diagnosis in this information, leading to hallucination. We observed that this issue becomes even more pronounced in the test set, likely due to a distribution shift toward more complex cases requiring a higher average number of tool calls and longer context. 

Beyond reasoning deficits, \textit{Output Parsing Errors} (19.2\%) stem from the model's failure to produce structured final answers (Figure~\ref{fig:failure-case-analysis-combined}(a)), a technical issue addressable via robust post-processing. Furthermore, \textit{Instruction Adherence Failures} (12.3\%) occur when the agent disregards formatting constraints (e.g., providing free-form text instead of selecting specific options, as in Figure~\ref{fig:failure-case-analysis-combined}(b)). We attribute the prevalence of both reasoning hallucinations and instruction failures primarily to the inherent limitations of TxAgent's compact architecture (base model Llama-3.1-8B~\cite{llama3.1} with 8B parameters), which struggles to maintain coherence and strict logic over long-horizon inference tasks. Recent research has examined the extent and timing of knowledge forgetting that occurs when large language models are fine-tuned on downstream tasks, revealing that domain-knowledge benchmarks such as MMLU can experience considerable performance drops after continual training on task-specific data~\cite{chen2020recall,kotha2023understanding,kalajdzievski2024scaling}. This raises a pivotal question: Can this performance gap be bridged simply by substituting the backbone with a more powerful foundation model? To investigate this, we conducted extensive benchmarking across a wide range of current mainstream open-source and closed-source models, as shown in Table~\ref{tab:opensource-model-comparison} and Table~\ref{tab:closedsource-model-comparison}.

\begin{table}[htbp]
  \centering
  \caption{Open-source Model Performance Comparison}
  \begin{tabular}{|l|c|c|}
    \hline
    \textbf{model name} & \textbf{phase1}\tablefootnote{phase1 refers to \texttt{testset\_phase1} (public leaderboard) and phase2 refers to \texttt{testset\_phase2} (private leaderboard). For details, see: \url{https://www.kaggle.com/competitions/cure-bench/data}} & \textbf{phase2} \\
    \hline
    TxAgent~\cite{gao2025txagent} & 65.214 & 69.325 \\
    \hline
    Llama-3.1-8B~\cite{llama3.1} & 27.912 & 29.715 \\
    \hline
    Llama3-Med42-8B~\cite{med42v2} & 46.125 & 50.407 \\
    \hline
    Qwen3-8B~\cite{qwen3technicalreport} & 28.134& 36.264 \\
    \hline
    Qwen-3-32B-Medical-Reasoning~\cite{qwen332bmedical} & 36.513 & 40.791 \\
    \hline
    Baichuan-M2~\cite{baichuan-m2} & 29.944 & 30.071 \\ 
    \hline
    medgemma-4b-it~\cite{medgemma} & 50.331 & 54.366 \\
    \hline
    gpt-oss-20b~\cite{openai2025gptoss120bgptoss20bmodel} & 45.425 & 47.644 \\
    \hline
    gpt-oss-120b~\cite{openai2025gptoss120bgptoss20bmodel} & 45.377 & 39.896 \\
    \hline
  \end{tabular}
  \label{tab:opensource-model-comparison}
\end{table}

Table~\ref{tab:opensource-model-comparison} reveals a stark contrast in performance between TxAgent and other open-source models. Despite the general capabilities of leading open-source foundation models like Llama-3.1-8B~\cite{llama3.1} and Qwen3-8B~\cite{qwen3technicalreport}, their zero-shot performance on CURE-Bench is significantly lower than that of TxAgent, with scores often falling below 30\%. Even specialized medical models such as Llama3-Med42~\cite{med42v2} and MedGemma~\cite{medgemma}, while outperforming general-purpose counterparts, still lag behind TxAgent by a substantial margin (approximately 15-20 points). This discrepancy underscores a critical insight: general-purpose reasoning or even broad medical knowledge is insufficient for tasks that require dynamic, real-time information access. Without fine-tuning on the specific syntax and semantics of the CURE-Bench toolset, standard open-source models struggle to effectively navigate the retrieval process, making them unable to access the ground-truth evidence required for accurate clinical decisions. This observation echoes the findings of ReAct~\cite{yao2022react,schick2023toolformer,qian2025toolrl}, where models significantly improve their tool-use capabilities through targeted training compared to their original, untrained versions.

\begin{table}[htbp]
  \centering
  \caption{Closed-source Model Performance Comparison}
  \begin{tabular}{|l|c|c|c|}
    \hline
    \textbf{model name} & \textbf{with tools} & \textbf{phase2} \\
    \hline
    gemini-2.5-flash & $\times$ & 63.104 \\
    \hline
    gemini-2.5-flash & \textit{search} & 69.627 \\ 
    \hline
    gemini-2.5-pro & $\times$ & 67.753 \\
    \hline
    gemini-2.5-pro & \textit{search} & 74.806 \\
    \hline
    gemini-3-pro-preview & $\times$ & 70.985 \\
    \hline
    gemini-3-pro-preview & \textit{search} & 81.283 \\
    \hline
  \end{tabular}
  \label{tab:closedsource-model-comparison}
\end{table}

Conversely, our evaluation of state-of-the-art closed-source models, presented in Table~\ref{tab:closedsource-model-comparison}, highlights a different limitation. Models with superior reasoning capabilities, such as the Gemini 2.5 series~\cite{google2025gemini25pro}, achieve respectable scores (60-75 range), particularly when augmented with generic search capabilities.\footnote{We also benchmarked the newly released Gemini 3 Pro (available post-competition) in Table~\ref{tab:closedsource-model-comparison} to gauge the trajectory of foundation models. Its impressive score of 81.283\% with search highlights the rapid advancement of reasoning engines. However, as this model was unavailable during the development phase, our architectural design and primary ablation studies focus on Gemini 2.5 era. It is promising to get much better result using Gemini 3 pro model.} 

A closer examination of the results shows that dynamic information retrieval plays a pivotal role in clinical reasoning tasks. With generic search enabled, both Gemini 2.5 Pro (74.8\%) and Gemini 2.5 Flash (69.6\%) exhibit clear performance gains, significantly surpassing their own baselines without search and matching or exceeding the performance of the task-specific TxAgent (69.3\%). For instance, Gemini 2.5 Pro without search (67.8\%) falls short of TxAgent despite its stronger reasoning abilities, but the addition of search alone yields a performance boost of nearly 7 points. This highlights that access to real-time, dynamic knowledge is essential even for large-scale foundation models.

However, there remains an open question: can a task-specialized agent like TxAgent, with its fine-grained tool-calling capabilities, further enhance information retrieval efficiency beyond what general-purpose search provides? In particular, as the reasoning capabilities of foundation models continue to advance rapidly, designing more precise, context-aware tool invocation mechanisms may be key to unlocking even greater downstream reasoning efficiency. Integrating the complementary strengths of dynamic search (for breadth and recency) and structured, task-specific tool execution (for precision) is likely to close the gap and maximize the utilization of available biomedical knowledge. Consequently, we propose a collaborative multi-agent framework that decouples these competencies. 

\begin{figure}[htbp]
  \centering
  \includegraphics[width=0.85\linewidth]{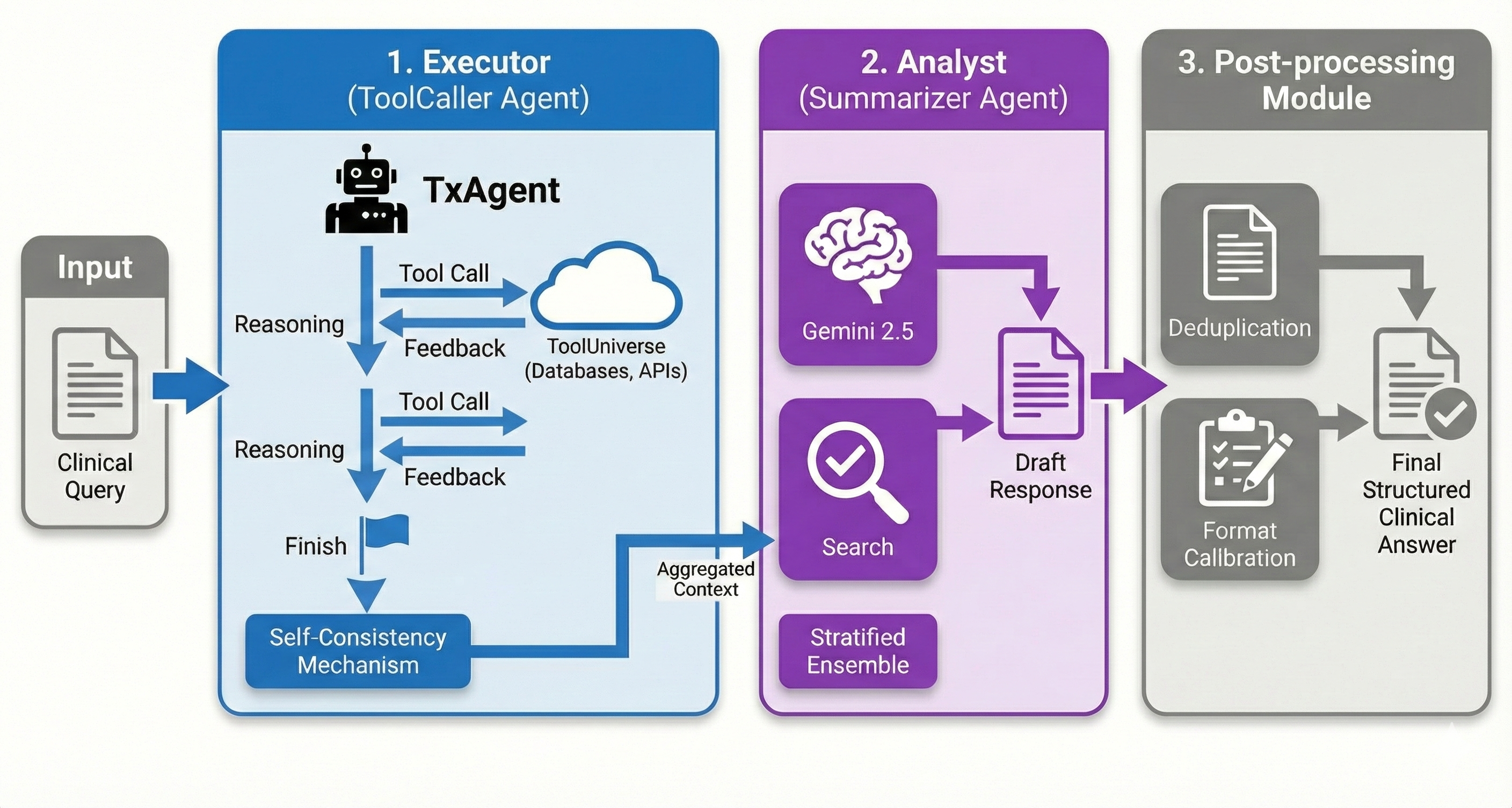}
  \caption{
    \textbf{Overview of the Executor-Analyst Collaborative Framework.} We address the context utilization failure by decoupling clinical reasoning into three specialized phases: 
    (1) \textbf{The Executor:} This agent focuses solely on precise information retrieval, utilizing a self-consistency mechanism to aggregate top-$k$ evidence and reasoning trace from the ToolUniverse.
    (2) \textbf{The Analyst:} Freed from tool-use syntax, this long-context foundation model acts as a reasoner, performing fact verification and supplementing tool-missing information via search and synthesis on the noisy evidence stream. Note that while this figure depicts a linear flow, we further enhance performance via a \textit{Stratified Ensemble} topology (detailed in Section~\ref{sec:topology}).
    (3) \textbf{Post-processing Module:} A deterministic layer ensuring format compliance and deduplication.
  }
  \label{fig:multi-agent-overview}
\end{figure}

Our approach bridges the context utilization failure gap by assigning TxAgent as a specialized Executor for accurate information retrieval and Gemini 2.5 as a clinical Analyst for evidence synthesis and diagnosis. Through the integration of search for fact-checking and supplementing tool-missing information, we enable robust, high-fidelity clinical decision-making without requiring costly domain-specific pre-training as illustrated in Figure~\ref{fig:multi-agent-overview}.

\section{Methods}

In this section, we propose the \textbf{Executor-Analyst Collaborative Framework}, a modular architecture that decouples the distinct competencies of precise tool execution (``the hand'') and high-level clinical reasoning (``the brain'') (Figure \ref{fig:multi-agent-overview}).

The framework decomposes the therapeutic reasoning process into multi-agent workflow, while a specialized agent responsible for navigating the ToolUNiverse to retrieve raw, unstructured evidence and a long-context foundation model that synthesizes noisy evidence into coherent clinical rationales. Then a deterministic layer ensuring deduplication and output format compliance. By separating these roles, we leverage the syntactic precision of fine-tuned smaller models for tool interactions while harnessing the robust reasoning capabilities of large-scale foundation models for evidence synthesis.

\subsection{The Executor: Specialized Tool Retrieval Agent}

The Executor serves as the system's interface with external biomedical knowledge. For this role, we utilize TxAgent, a Llama-3.1-8B model fine-tuned specifically for therapeutic reasoning. Unlike general-purpose models, the Executor is optimized to interact with ToolUniverse, a comprehensive library of over 200 validated biomedical tools (including FDA labels, Open Targets, and Human Phenotype Ontology).

The Executor does not generate the final clinical answer. Instead, its objective is context Aggregation. It parses the input clinical scenario, decomposes it into sub-queries, and orchestrates multi-step tool invocations and reasoning. To mitigate the stochasticity of single-shot retrieval, we implement a self-consisteny mechanism~\cite{wang2022selfconsistency} by instantiating multiple parallel Executor agents ($n_1$) for a single query. Rather than relying on a single reasoning path, we aggregate the tool outputs from these parallel executions, retaining the top-$k$ most frequently selected tool calls and the most frequent choice reasoning trace. This ensures that the downstream Analyst receives a comprehensive and robust set of evidence, minimizing the risk of missing critical contraindications or drug interactions due to retrieval errors.

\subsection{The Analyst: Long-Context Clinical Reasoner}

To address the reasoning deficit inherent in smaller, tool optimized models, the Analyst employs Gemini 2.5 variant as a dedicated reasoning backbone. Ingesting the Aggregated Context, a noisy, unstructured stream of raw observations from the Executor, this agent is relieved of the syntactic burden of tool invocation and multi-turn dialogue. Instead, it concentrates exclusively on high-level clinical processing: cross-referencing retrieved tool outputs and performing supplementary internet searches against the patient's specific comorbidities.

Leveraging its vast context window and superior ``System 2'' capabilities, the Analyst synthesizes this verified information into a coherent chain-of-thought rationale. It effectively filters irrelevant noise, actively searches the internet when tool-retrieved evidence is insufficient, and resolves conflicting data points without hallucinating non-existent medical facts. This rigorous process culminates in Clinical Synthesis, producing a preliminary draft that firmly grounds the diagnosis and supporting arguments in verified evidence.

\subsection{Architecture Topology: Global Pooling vs. Stratified Ensemble}
\label{sec:topology}

\begin{figure}[htbp]
  \centering
  \begin{subfigure}[b]{0.4\linewidth}
    \centering
    \includegraphics[width=0.99\linewidth]{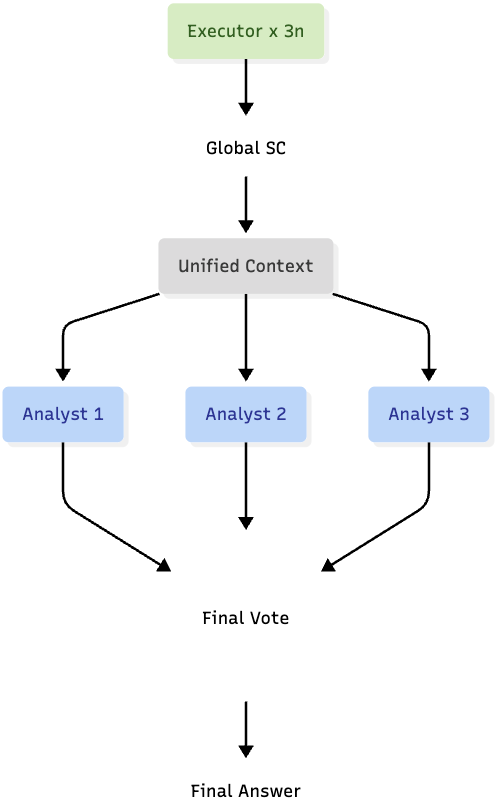}
    \caption{Config A: Global Pooling (Early Fusion)}
    \label{fig:config-a}
  \end{subfigure}
  \hfill
  \begin{subfigure}[b]{0.47\linewidth}
    \centering
    \includegraphics[width=0.99\linewidth]{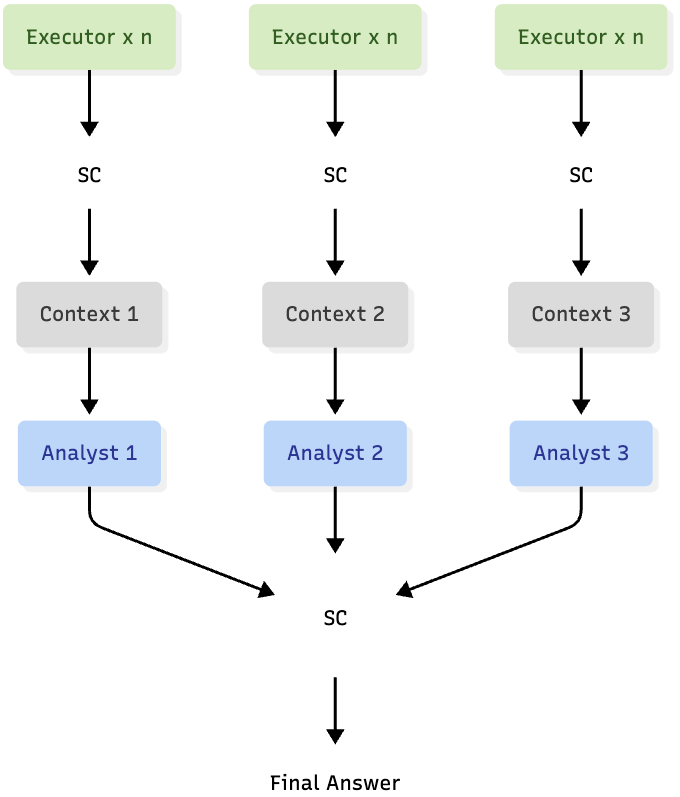}
    \caption{Config B: Stratified Ensemble (Late Fusion)}
    \label{fig:config-b}
  \end{subfigure}
  \caption{
    \textbf{Comparison of Topological Configurations.} 
    (a) \textbf{Global Pooling} aggregates all evidence into a single context. While reducing noise, it risks filtering out rare but valid cues. 
    (b) \textbf{Stratified Ensemble} (Ours) partitions agents into parallel subgroups. Late Fusion approach preserves diverse reasoning paths until the final stage, mitigating the information bottleneck.
  }
  \label{fig:multi-agent-configs}
\end{figure}

A critical factor in multi-agent orchestration is balancing consensus against diversity. We investigate two distinct topological configurations to optimize the collaboration between Executors and Analysts using a fixed computational budget (Figure~\ref{fig:multi-agent-configs}).

\textbf{Configuration A: Global Pooling (Early Fusion).} In this baseline configuration (Figure~\ref{fig:multi-agent-configs}(a)), we maximize the retrieval sample size by pooling all Executor instances ($N_{total}$) into a single aggregation step. Then the generated context will be processed by $n_2$ independent Analyst agents and perform another self consistency. 

While this approach effectively filters out retrieval noise, it introduces a severe information bottleneck: by forcing a consensus too early in the pipeline, minority yet critical evidence is often discarded before reaching the reasoning stage.

\textbf{Configuration B: Stratified Ensemble (Late Fusion).} To address the bottleneck, we propose a \textbf{Stratified Ensemble} strategy (Figure~\ref{fig:multi-agent-configs}(b)) that preserves information diversity. Instead of a monolithic pool, we partition the Executor budget into $n_2$ parallel subgroups, each subgroup is with $n_1=N_{total}/n_2$ Executor instances.
\begin{enumerate}
    \item \textbf{Parallel Pipelines:} Each subgroup independently aggregates its retrieval results, forming distinct context streams tailored to different retrieval paths.
    \item \textbf{Independent Reasoning:} Each unique context is ingested by a dedicated Analyst, creating fully independent end-to-end reasoning chains.
    \item \textbf{Late Fusion:} The final decision is reached via self-consisteny mechanism on the Analyst's \textit{final answers}, rather than the intermediate evidence.
\end{enumerate}

This ensures that diverse retrieval paths are maintained throughout the reasoning process, significantly reducing collective hallucination and improving robustness on complex, ambiguous cases.

\subsection{Post-processing Module: Deterministic Output Calibration} 

The final stage addresses output parsing errors and stochastic inconsistencies. The post-processing module is a lightweight, deterministic component that accepts the Analyst's draft. It performs two key functions: 1) Format Calibration, which employs regular expressions (Regex) to map natural language conclusions into the strict output schema required by the benchmark; and 2) Response Deduplication, which enforces internal consistency across identical queries. By standardizing outputs for recurrent inputs, this layer mitigates the inherent generation variance of LLMs, ensuring the system exhibits the deterministic behavior critical for reliable clinical decision support.

\section{Experiments and Ablation studies}

\textbf{Dataset.} Due to the late release of the \texttt{phase2} set (phase2 testset), we prioritize \texttt{phase2} for multi-agent structure selection and final model comparisons, while retaining \texttt{phase1} for ablation studies where resource constraints prevented re-evaluation. \footnote{Implementation details, such as temperature could be found in https://github.com/June01/CureAgent.}

\subsection{Executor Calibration}

We first investigate the impact of generation diversity on the Executor's retrieval capabilities. We conducted a sweep of the temperature parameter on the TxAgent on \texttt{phase1} in Table~\ref{tab:temperature-results}, while performance peaks at a temperature of $0.8$, achieving a score of $65.214$, indicating that a moderate increase in generation randomness helps TxAgent discover better tool call strategies. However, setting the temperature too high ($0.9$) leads to performance degradation, likely due to excessive output variability. These findings suggest careful calibration of the temperature hyperparameter is crucial for balancing exploration and reliability in clinical tool-using agents. Consequently, we fix $T=0.8$ for subsequent experiments.

\begin{table}[htbp]
  \centering
  \caption{Performance of TxAgent with Different Temperature Settings on \texttt{phase1}.}
  \begin{tabular}{|c|c|c|}
    \hline
    \textbf{model} & \textbf{temperature} & \textbf{phase1} \\
    \hline
    txagent & 0.6 & 58.950 \\ \hline
    txagent & 0.7 & 59.248 \\ \hline
    txagent & \textbf{0.8} & \textbf{65.214} \\ \hline  
    txagent & 0.9 & 56.747 \\
    \hline
  \end{tabular}
  \label{tab:temperature-results}
\end{table}

\subsection{Self-consistency Mechanism}

\begin{figure}[htbp]
  \centering
  \includegraphics[width=0.5\linewidth]{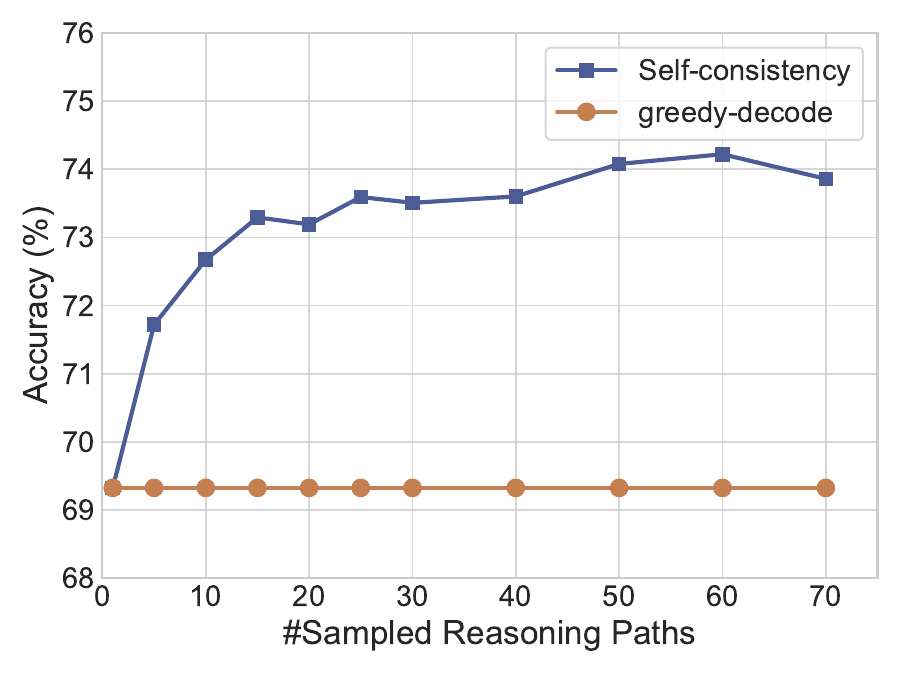}
  \caption{Performance of TxAgent with Self-consistency Mechanisms on phase2.}
  \label{fig:self-consistency-results}
\end{figure}

We evaluate the Self-Consistency (SC) mechanism~\cite{wang2022selfconsistency} for enhancing the Executor's retrieval performance on \texttt{phase2}. As shown in Figure~\ref{fig:self-consistency-results}, increasing the number of sampled reasoning paths ($n$) and aggregating their outputs via majority voting leads to notable accuracy improvements over standard greedy decoding (orange line), which remains fixed at $69.3\%$. The SC method (blue line) achieves rapid gains in the low-sample regime ($n < 15$), with accuracy climbing to about $73.3\%$. Beyond $n > 20$, the improvement gradually plateaus, reaching approximately $74.2\%$ at $n=60$. 

\subsection{Collaborative Architecture Effectiveness}

\begin{table}[htbp]
  \centering
  \caption{Main results on \texttt{phase2}. Comparison of Single-Agent vs. Collaborative Architectures. $n_1$: Number of Executors. $n_2$: Number of Analysts. \textbf{Config A}: Global Pooling. \textbf{Config B}: Stratified Ensemble.}
  \begin{tabular}{|c|c|c|c|c|}
    \hline
    \textbf{Agent1} & $n_1$ & \textbf{Agent2} & $n_2$ & \textbf{phase2} \\
    \hline
    \multicolumn{5}{|c|}{\textit{Baselines}} \\ \hline
    gemini-2.5-flash & 1 & - & - & 63.104\\ \hline
    TxAgent & 1 & - & - & 69.325 \\ \hline
    \multicolumn{5}{|c|}{\textit{Single Agent with Self-Consistency}} \\ \hline
    gemini-2.5-flash & 3 & - & - & 65.678\\ \hline
    TxAgent & 10 & - & - & 72.673 \\ \hline
    TxAgent & 30 & - & - & 73.508 \\ \hline
    \multicolumn{5}{|c|}{\textit{Collaborative}} \\ \hline
    TxAgent (Config A) & 1 & gemini-2.5-flash & 1 & 74.701 \\ \hline
    TxAgent (Config A) & 10 & gemini-2.5-flash & 1 & 78.527 \\ \hline
    TxAgent (Config A) & 30 & gemini-2.5-flash & 1 & 79.311 \\ \hline
    TxAgent (Config A) & 30 & gemini-2.5-flash & 3 & 80.510 \\ \hline
    TxAgent (Config B) & 10 & gemini-2.5-flash & 3 & 81.367 \\ \hline
    TxAgent (Config B) & 10 & gemini-2.5-flash (with search) & 3 & 83.803 \\ 
    \hline
  \end{tabular}
  \label{tab:multi-agent-results}
\end{table}

We proceed to comprehensively evaluate the full Executor-Analyst Framework, in which Self-Consistency (SC) mechanisms are incorporated into both the retrieval and reasoning components. This assessment examines the progression from traditional single-agent baselines to our advanced Stratified Ensemble architecture. As demonstrated in Table~\ref{tab:multi-agent-results}, the results robustly support the validity of our design principles:

1) Decoupling Gains: The most significant jump occurs when combining a single Executor with a single Analyst ($74.701\%$), suppressing the isolated TxAgent baseline ($69.325\%$) and the standalone Gemini baseline ($63.104\%$). This confirms that decoupling precise tool execution from high-level reasoning is the primary driver of performance.

2) Topology Matters (Early vs. Late Fusion): While scaling the Executor pool to $N_{total}=10 \times 3$, i.e., $n_1=30$ and $n_2=3$ in Config A yields strong results ($80.51\%$), it still faces an information bottleneck. By switching to Config B, which employs a Stratified Ensemble, a late fusion strategy with parallel pipelines ($N_{total}=10 \times 3$), i.e., $n_1=10$ and $n_2=3$ , we achieve the state-of-the-art performance of $81.367\%$. This verifies that preserving diverse retrieval contexts prevents the premature filtering of critical evidence, effectively reducing collective hallucination. 

Note that the previous results are using Gemini 2.5 Flash without search; when equipped with search capabilities, Gemini 2.5 Flash achieves $83.803\%$ accuracy, which is the configuration we used for the leaderboard results.

\section{Open Challenges and Conclusion}


Our investigation into the Executor-Analyst framework has validated the efficacy of decoupling tool execution from clinical reasoning. However, we also found several critical challenges emerge, thus we outline the roadmap for the next generation of clinical agents.

\subsection{Scalability in Massive Action Spaces}
\label{sec:future-tool-scale}

A primary limitation of current tool-use agents is the degradation of retrieval accuracy as the tool library expands. To quantify this, we conducted a preliminary stress test by upgrading ToolUniverse from v1.0 (200+ tools) to v2.0 (600+ distinct tools)~\cite{gao2025tooluniverse}. As shown in Table~\ref{tab:tooluniverse}, despite the increased domain coverage, the model's accuracy decreased from 92.0\% to 87.5\%. 

\begin{table}[htbp]
  \centering
  \caption{Performance of TxAgent on the Validation Set with Different ToolUniverse Versions.}
  \begin{tabular}{|c|c|c|}
    \hline
    \textbf{model} & \textbf{ToolUniverse 1.0 (200+ tools)} & \textbf{ToolUniverse 2.0 (600+ tools)} \\
    \hline
    TxAgent & 92.0\% & 87.5\% \\
    \hline
  \end{tabular}
  \label{tab:tooluniverse}
\end{table}

This performance drop highlights the \textit{Curse of Dimensionality} in tool retrieval. Future work must address this by moving beyond flat retrieval spaces. We plan to investigate: 1) \textbf{Hierarchical Tool Indexing:} Grouping tools into clinical taxonomies to allow coarse-to-fine retrieval. 2) \textbf{Training-free Adaptation:} Exploring methods to teach models to navigate massive toolsets ($>1000$) via In-Context Learning (ICL) or RAG-based documentation retrieval, eliminating the cost of fine-tuning for every new API update. 3) \textbf{Curriculum Learning:} Progressively exposing the Executor to larger tool subsets during training to improve robustness against distractors.

\subsection{The Context-Performance Paradox and Token Efficiency}
\label{sec:future-context}

\begin{figure}[htbp]
  \centering
  \begin{subfigure}{0.48\linewidth}
    \includegraphics[width=\linewidth]{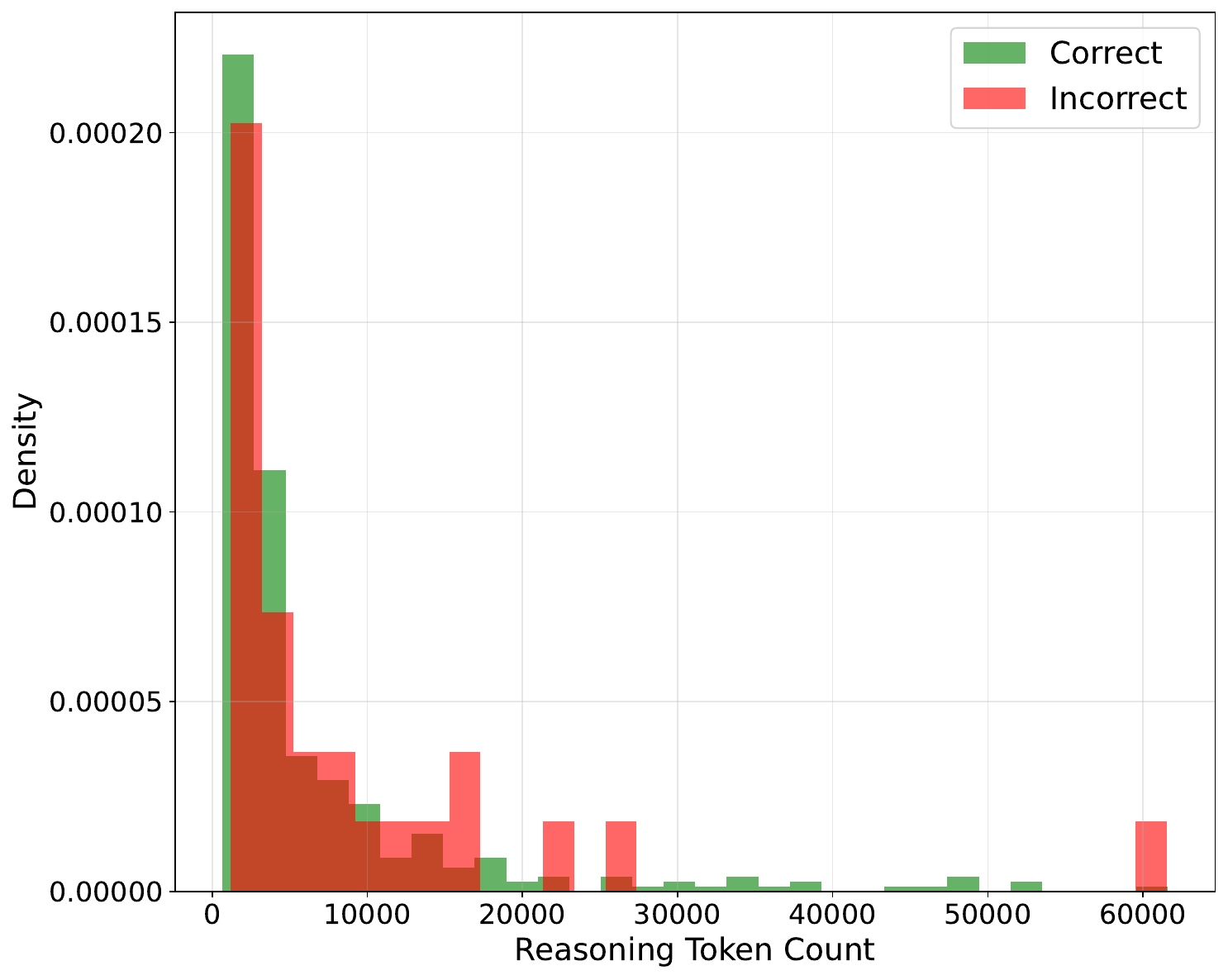}
    \caption{Reasoning token distribution for samples.}
    \label{fig:token-distribution}
  \end{subfigure}
  \hfill
  \begin{subfigure}{0.48\linewidth}
    \includegraphics[width=\linewidth]{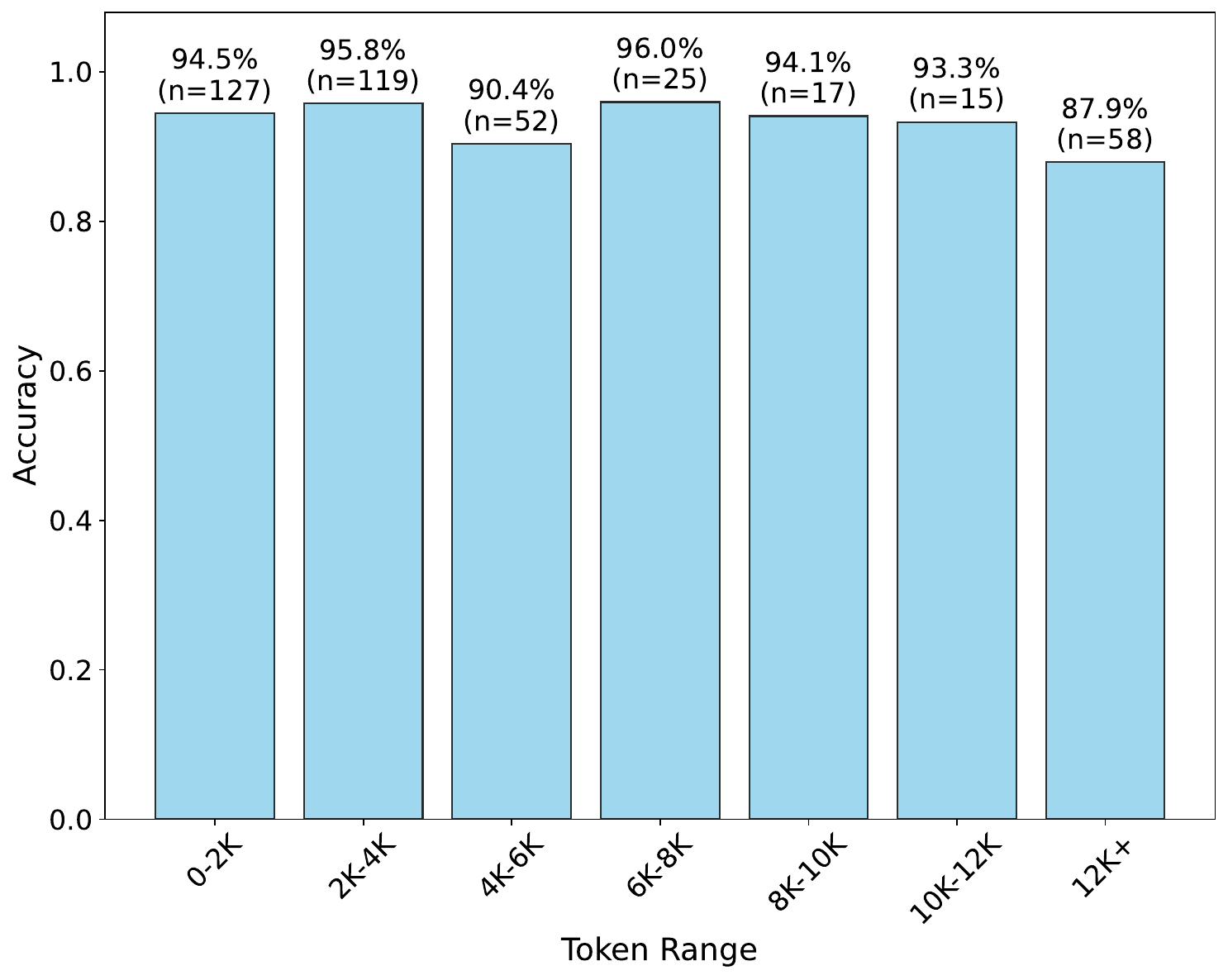}
    \caption{Accuracy across different reasoning token ranges.}
    \label{fig:token-accuracy}
  \end{subfigure}
  \caption{Analysis of performance vs context length.}
  \label{fig:token-length-analysis}
\end{figure}

Besides, our analysis reveals a counter-intuitive \textit{Context-Performance Paradox}: extending the TxAgent's context window does not linearly correlate with better reasoning. As visualized in Figure~\ref{fig:token-length-analysis}, we observe a sharp performance decline (from 94\% to 87.93\%) when reasoning contexts exceed 12,000 tokens. The long-tail distribution of incorrect samples suggests that excessive raw evidence introduces noise that overwhelms the model's attention mechanism, echoing recent findings by Du et al.~\cite{du2025context}.

To mitigate this, future iterations will focus on information compression and early rejection. Instead of feeding all aggregated evidence to the agent, we aim to implement confidence-based filtering strategies similar to DeepConf~\cite{fu2025deepconf}. By estimating the semantic redundancy of tool outputs and rejecting irrelevant evidence before it consumes the context window, we can improve the signal-to-noise ratio and significantly reduce inference costs.

\subsection{Evolving the Reasoning Engine}

The landscape of foundation models is evolving rapidly. While our current framework relies on Gemini 2.5 variants, the emergence of next-generation models (e.g., Gemini 3 Pro) offers new possibilities for the Analyst role. Our preliminary zero-shot benchmarks (Table~\ref{tab:closedsource-model-comparison}) indicate that Gemini 3 Pro already achieves superior performance. Future work will explore shifting the boundary between the Executor and Analyst. As foundation models become more capable of zero-shot tool use, the specialized Executor might evolve from a fine-tuned model into a lightweight prompt-engineered module. However, the core principle of our architecture—\textit{decoupling execution from reasoning}—remains vital. This modularity allows our framework to seamlessly integrate stronger backbones without the need for expensive end-to-end retraining, ensuring adaptability in a fast-paced ecosystem.

\section{Conclusion}

We address the limitations of monolithic medical agents by introducing the Executor-Analyst collaborative framework, which decouples tool manipulation (the hand) from clinical synthesis (the brain) to resolve the conflict between syntactic precision and semantic reasoning. Ablation studies on CURE-Bench reveal that reliable decision-making requires not just tool access but a topological structure preserving information diversity—our Stratified Ensemble (Late Fusion) mitigates early information bottlenecks and handles diverse retrieval paths. Beyond immediate gains, training-free architectural engineering enables rapid adaptation: future models can upgrade Analyst or expand the Executor's ToolUniverse without retraining. Remaining open questions include the \textit{Context-Performance Paradox} (noise from longer contexts) and the \textit{Curse of Dimensionality} in large action spaces. By emphasizing interaction design (e.g., early rejection, hierarchical orchestration) over parameter updates, our work provides a robust and future-proof preview toward scalable, trustworthy clinical AI.

\enlargethispage{-\baselineskip} 

\section*{Acknowledgments}

All experiments ran on one AMD MI300X GPU, thanks to AMD for hardware support. Thanks also to Google AI Developer Programs team and GDE Healthcare Circle for Google Cloud and Gemini credits. We also gratefully acknowledge Synapnote.ai for the support.


\bibliographystyle{unsrtnat}
\bibliography{sample}

\end{document}